\documentclass[sigconf]{aamas}

\usepackage{multirow}

\usepackage[ruled,vlined]{algorithm2e}
\usepackage{balance} 

\makeatletter
\gdef\@copyrightpermission{
  \begin{minipage}{0.2\columnwidth}
   \href{https://creativecommons.org/licenses/by/4.0/}{\includegraphics[width=0.90\textwidth]{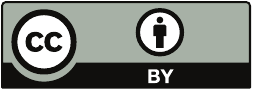}}
  \end{minipage}\hfill
  \begin{minipage}{0.8\columnwidth}
   \href{https://creativecommons.org/licenses/by/4.0/}{This work is licensed under a Creative Commons Attribution International 4.0 License.}
  \end{minipage}
  \vspace{5pt}
}
\makeatother

\setcopyright{ifaamas}
\acmConference[AAMAS '25]{Proc.\@ of the 24th International Conference
on Autonomous Agents and Multiagent Systems (AAMAS 2025)}{May 19 -- 23, 2025}
{Detroit, Michigan, USA}{Y.~Vorobeychik, S.~Das, A.~Nowé  (eds.)}
\copyrightyear{2025}
\acmYear{2025}
\acmDOI{}
\acmPrice{}
\acmISBN{}

\acmSubmissionID{1020}

\title[AAMAS-2025 Formatting Instructions]{Feature Engineering for Agents: An Adaptive Cognitive Architecture for Interpretable ML Monitoring}

\author{Gusseppe Bravo-Rocca}
\orcid{0000-0001-6824-1124}
\email{gusseppe.bravo@bsc.es}
\affiliation{%
  \institution{Barcelona Supercomputing Center}
  \city{Barcelona}
  \country{Spain}
}
\author{Peini Liu}
\orcid{0000-0003-0058-8732}
\email{peini.liu@bsc.es}
\affiliation{%
  \institution{Barcelona Supercomputing Center}
  \city{Barcelona}
  \country{Spain}
}
\author{Jordi Guitart}
\orcid{0000-0003-0751-3100}
\email{jordi.guitart@bsc.es}
\affiliation{%
  \institution{Barcelona Supercomputing Center\\Universitat Politècnica de Catalunya}
  \city{Barcelona}
  \country{Spain}
}
\author{Rodrigo M Carrillo-Larco}
\orcid{0000-0002-2090-1856}
\email{rmcarri@emory.edu}
\affiliation{%
  \institution{Emory University}
  \city{Atlanta}
  \state{GA}
  \country{USA}
}
\author{Ajay Dholakia}
\orcid{0009-0007-8973-6063}
\email{adholakia@lenovo.com}
\affiliation{%
  \institution{Lenovo Infrastructure Solutions Group}
  \city{Morrisville}
  \state{NC}
  \country{USA}
}
\author{David Ellison}
\orcid{0000-0002-0752-5569}
\email{dellison@lenovo.com}
\affiliation{%
  \institution{Lenovo Infrastructure Solutions Group}
  \city{Morrisville}
  \state{NC}
  \country{USA}
}

\begin{abstract}
Monitoring Machine Learning (ML) models in production environments is crucial, yet traditional approaches often yield verbose, low-interpretability outputs that hinder effective decision-making. We propose a cognitive architecture for ML monitoring that applies feature engineering principles to agents based on Large Language Models (LLMs), significantly enhancing the interpretability of monitoring outputs. Central to our approach is a Decision Procedure module that simulates feature engineering through three key steps: Refactor, Break Down, and Compile. The Refactor step improves data representation to better capture feature semantics, allowing the LLM to focus on salient aspects of the monitoring data while reducing noise and irrelevant information. Break Down decomposes complex information for detailed analysis, and Compile integrates sub-insights into clear, interpretable outputs. This process leads to a more deterministic planning approach, reducing dependence on LLM-generated planning, which can sometimes be inconsistent and overly general.
The combination of feature engineering-driven  planning and selective LLM utilization results in a robust decision support system, capable of providing highly interpretable and actionable insights. Experiments using multiple LLMs demonstrate the efficacy of our approach, achieving significantly higher accuracy compared to various baselines across several domains.

\end{abstract}


\newcommand{\BibTeX}{\rm B\kern-.05em{\sc i\kern-.025em b}\kern-.08em\TeX}

\begin{document}


\pagestyle{fancy}
\fancyhead{}


\maketitle 


\section{Introduction}

\begin{figure}[t]
\centering
\includegraphics[width=0.8\columnwidth]{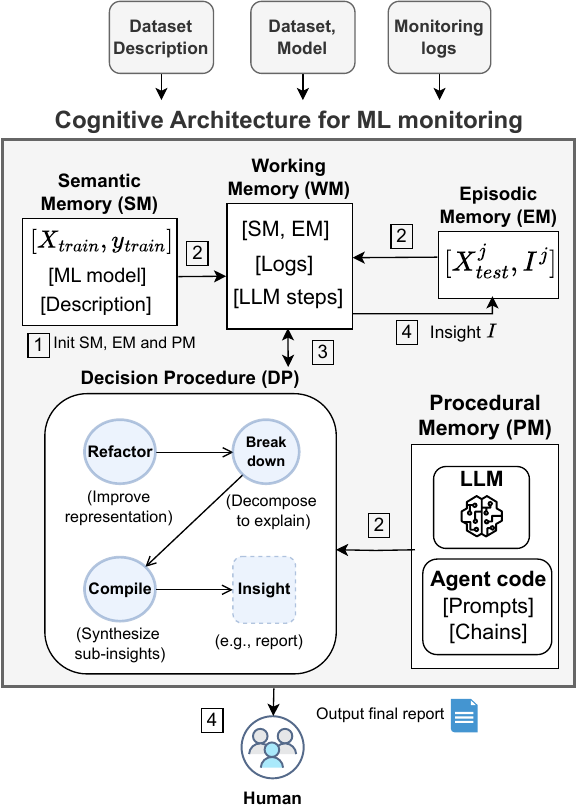}
\caption{Cognitive architecture for ML monitoring. The system integrates Procedural (PM), Episodic (EM), Semantic (SM), and Working (WM) Memory. The central Decision Procedure (DP) implements a feature engineering-inspired approach: Refactor, Break Down, and Compile. SM stores reference data $[X_{\text{train}}, y_{\text{train}}]$, the ML model, and dataset descriptions. EM captures current test data $[X_{\text{test}}^j, I^j]$ and insights. WM facilitates information exchange. PM sets the LLM and agent code. This architecture enhances monitoring output interpretability, providing clear, actionable insights for human operators.}
\label{fig:cognitive_architecture}
\end{figure}

Monitoring Machine Learning (ML) models in production environments is a critical task, as model performance can degrade over time due to various factors \cite{monitoring_ml_models, eck2022monitoring}. Traditional monitoring approaches, such as distribution drift detection and feature attribution \cite{rabanser2019failing, lipton2018detecting}, while important, often require substantial technical expertise to interpret and act upon. As the number of deployed models increases, the time required for thorough analysis of each model's performance becomes a significant bottleneck in maintaining up-to-date and reliable ML systems.
A pressing question emerges: can we leverage Large Language Models (LLMs) to automate and enhance the monitoring of ML models? Specifically, can LLMs analyze the outputs of monitoring tools, interpret these results, and relate them to the dataset to provide meaningful, actionable insights? This capability would enable the generation of detailed, interpretable reports on model issues, facilitating crucial decisions such as model retraining, new data labeling, or model replacement.
While established monitoring tools like Alibi Detect\footnote{https://github.com/SeldonIO/alibi-detect} provide valuable technical metrics (e.g., drift scores, SHAP values), interpreting these outputs often requires significant expertise. There is an opportunity to complement them by making their outputs more interpretable and actionable through natural language processing.
To address this challenge, we propose a novel cognitive architecture for ML monitoring that applies feature engineering principles to LLM-based agents, significantly enhancing the interpretability of monitoring outputs. Our approach, termed CAMA (Cognitive Architecture for Monitoring Agent), combines structured memory components with a sophisticated decision procedure to generate highly interpretable and actionable insights (see Figure \ref{fig:cognitive_architecture}).

Central to our architecture is the Decision Procedure (DP) module, which implements a feature engineering-inspired approach through three key steps:
\begin{itemize}
\item Refactor: Improves data representation to better capture feature semantics, allowing the LLM to focus on salient aspects of the monitoring data while reducing noise.
\item Break Down: Decomposes complex information for detailed analysis, enabling comprehensive understanding of individual features and their interactions.
\item Compile: Integrates sub-insights into clear, interpretable outputs, providing a holistic view of the model's performance.
\end{itemize}
This process leads to a more deterministic planning approach, reducing dependence on LLM-generated planning, which can sometimes be inconsistent and overly general. Our architecture incorporates principles from cognitive science, particularly the structured memory and thinking processes inspired by human cognition \cite{cognitive_architectures}. This approach ensures efficiency while maintaining the capability for deep, thoughtful analysis when required.
The effectiveness of our approach is demonstrated through experimental evaluations using four different LLMs, ranging from smaller models like llama-3.2-1b to large-scale models such as llama3-70b \cite{touvron2023llama} and gpt-4o-mini. Our method, CAMA, consistently outperforms various baselines across different metrics. 
In our experiments, we work with three distinct datasets with different level of complexity designed to simulate real-world distribution drift scenarios. These datasets represent diverse domains in financial services and customer's behavior prediction, allowing us to demonstrate the versatility and robustness of our approach in handling various types of distribution shifts commonly encountered in production environments.

Our contributions are the following:
\begin{itemize}
\item The development of a cognitive architecture for monitoring ML models that enhances the interpretability and actionability of monitoring reports.
\item A novel decision procedure that applies feature engineering principles to LLM-based analysis, resulting in more deterministic and reliable planning.
\item A comprehensive evaluation demonstrating the superiority of our approach across different LLM models and metrics, using datasets that mimic real-world distribution drift challenges.
\end{itemize}


\section{Related Work}
Various prompting and planning techniques have been investigated to enhance the decision-making capabilities of LLMs. This section covers key methodologies relevant to our approach, highlighting their strengths and limitations, and contrasting them with our adaptive cognitive architecture.
\subsection{Direct Prompting Techniques}

\textbf{Prompting (Standard):} This method presents the model with a specific question or task (bare prompt) and requests a response without additional context. While effective for simple queries, it often falls short in complex reasoning tasks \cite{brown2020language, wei2022finetuned}. For instance, while a model might easily answer "What is the capital of France?", solving equations typically requires careful regularization or model selection. Our solution leverages a multi-faceted cognitive architecture that scales with task complexity, providing more accurate and contextually sensitive responses.

\textbf{Chain of Thought (CoT)} prompting \cite{wei2022chain, kojima2022large} enables LLMs to formulate their own "thinking procedure" by eliciting intermediate reasoning steps. This technique significantly improves performance on complex reasoning tasks. However, CoT lacks a structured mechanism to handle diverse and evolving datasets systematically. Our approach incorporates similar reflective processes within a cognitive architecture, allowing the model to focus on minimal context stored in Working Memory (WM) and avoid noise.

\textbf{Reflection} methods leverage the model's ability to reconsider and refine its previous responses, improving accuracy through iterative self-correction \cite{shinn2023reflexion}. While effective, Reflection is limited by its reliance on continuous feedback loops, which can be computationally expensive. Our approach mitigates these limitations by utilizing a structured Decision Procedure to balance quick assessments with thorough analyses, thereby optimizing resource utilization.
\subsection{Agentic Behavior Techniques}

\textbf{ReAct} \cite{yao2022react} combines reasoning and action at every step, excelling in dynamic, context-specific responses. However, it can rush to solutions without sufficient analysis when deeper understanding is required. In contrast, our approach embeds action-oriented strategies within a structured cognitive framework. This allows for dynamic responses while also ensuring deeper, more systematic analysis when needed. By applying feature engineering principles, our approach refines and compiles monitoring data, delivering more contextually relevant and actionable insights, balancing immediate action with thorough analysis.

\textbf{Self-Discover} methodology \cite{zhou2024selfdiscoverlargelanguagemodels} enables LLMs to autonomously generate reasoning structures, reducing dependency on pre-constructed prompts. While effective for novel problems, it may struggle with consistency across different contexts without a structured memory system. Our approach enhances self-discovery by using structured Semantic Memory (SM), allowing the LLM to leverage current data like dataset descriptions to improve reasoning and maintain contextually relevant insights.

The \textbf{Plan-and-Solve} technique \cite{wang2023plan} prompts models to plan a solution pathway before execution, improving zero-shot chain-of-thought reasoning. However, its linear approach can be restrictive for complex problem-solving. In contrast, our approach embeds planning in the architecture. This allows for a more flexible and dynamic approach, ensuring more efficient and accurate execution of complex reasoning tasks.

More recently, \textbf{PH-LLM} \cite{cosentino2024personal}, a fine-tuned variant of the Gemini model for interpreting personal health data, highlights the effectiveness of using domain-specific data to generate personalized insights. However, PH-LLM relies on model fine-tuning, whereas our approach is fully LLM-agnostic. Rather than customizing specific models, we employ structured memory and reasoning techniques that work across any LLM, regardless of size or architecture.

In contrast to existing methods, our approach integrates structured memory, feature engineering, and adaptive decision-making. This combination enables more efficient, accurate, and contextually relevant analysis of ML model performance, overcoming the limitations of current approaches in handling complex, evolving datasets in production environments.


\section{Approach}
While our architecture incorporates established memory components, our key innovation lies in the decision-making process (Refactor, Break Down, Compile) that applies feature engineering principles to enhance monitoring interpretability. CAMA operates primarily on the outputs of monitoring tools (e.g., drift metrics, SHAP values) rather than directly on raw datasets. The architecture interprets these monitoring results to provide actionable recommendations, such as model retraining timing or data pipeline adjustments. This design choice makes CAMA format-agnostic and applicable across different monitoring scenarios and metrics, regardless of the underlying data domains.

\subsection{Memory Modules}
Our architecture utilizes four distinct memory modules, each serving a specific role in the monitoring process:

\textbf{Procedural Memory} ($\mathcal{M_P}$): Stores the agent code, including prompts and chains, enabling effective utilization of the LLM within the cognitive architecture. It encompasses both explicit procedural knowledge encoded in the agent's code and implicit knowledge embedded in the LLM weights.

\textbf{Episodic Memory} ($\mathcal{M_E}$): Retains specific past instances of model monitoring, including test data $X_{\text{test}}^j$ and generated insights $I^j$. Formally represented as:
\begin{equation}
\mathcal{M_E} = {(X_{\text{test}}^j, I^j)}_{j=1}^n
\end{equation}
This memory is crucial for learning from historical data and enhancing decision-making based on past experiences.

\textbf{Semantic Memory} ($\mathcal{M_S}$): Contains generalized knowledge extracted from training data, the ML model, and monitoring tools. Defined as:
\begin{equation}
\mathcal{M_S} = ({[X_{\text{train}}, y_{\text{train}}], \mathcal{H}, \mathcal{T}})
\end{equation}
where $\mathcal{H}$ represents the ML model and $\mathcal{T}$ represents the tools. This memory supports contextual understanding and retrieval of relevant information.

\textbf{Working Memory} ($\mathcal{M_W}$): Holds the current context, including test data, intermediate insights, and ongoing reasoning processes (LLM steps). Formulated as:
\begin{equation}
\mathcal{M_W} = ({\mathcal{M_E}, \mathcal{M_S}, \text{LLM steps}})
\end{equation}
$\mathcal{M_W}$ facilitates dynamic adaptation of responses based on real-time analysis.
\subsection{Decision Procedure}
Our decision procedure is the core of the agent, it implements a feature engineering-inspired approach through three key steps: Refactor, Break Down, and Compile. This process enhances the interpretability and actionability of monitoring insights. Algorithm \ref{algo:decision_procedure} outlines the overall decision procedure.

\begin{algorithm} \SetAlgoLined \DontPrintSemicolon \textbf{Input}: Procedural Memory $\mathcal{M_P}$, Semantic Memory $\mathcal{M_S}$, Episodic Memory $\mathcal{M_E}$, Test data $X_{\text{test}}$; \BlankLine \textbf{Output}: Monitoring Report or Deep Insight $I_{\text{deep}}$; \BlankLine \textbf{Initialize Working Memory}: $\mathcal{M_W} \gets {\mathcal{M_E}, \mathcal{M_S}, \text{LLM steps}}$; \BlankLine \textbf{Refactor Step}: \Begin{ $I_{\text{ref}} \gets \text{Refactor}(X_{\text{test}}, \mathcal{M_S}, \mathcal{M_E})$; } \BlankLine \textbf{Break Down Step}: \Begin{ \ForAll{features $f_i \in I_{\text{ref}}$ \textbf{in parallel}}{ Generate prompt $p_i$ using $\mathcal{M_P}$ and $\mathcal{M_W}$; $r_{f_i} \gets \text{AnalyzeFeature}(f_i, p_i)$; } $I_{\text{div}} \gets {r_{f_i} \mid f_i \in I_{\text{ref}}}$; } \BlankLine \textbf{Compile Step}: \Begin{ $I_{\text{deep}} \gets \text{CompileReport}(I_{\text{div}}, \mathcal{M_P})$; } \BlankLine \textbf{Update Episodic Memory}: $\mathcal{M_E} \gets \mathcal{M_E} \cup {(X_{\text{test}}, I_{\text{deep}})}$; \BlankLine \Return $I_{\text{deep}}$; \caption{Decision Procedure} \label{algo:decision_procedure} \end{algorithm}

\subsubsection{Decision Steps}
The decision procedure consists of three key steps that work together to provide comprehensive monitoring analysis:

\begin{figure}[h]
\centering
\includegraphics[width=0.7\linewidth]{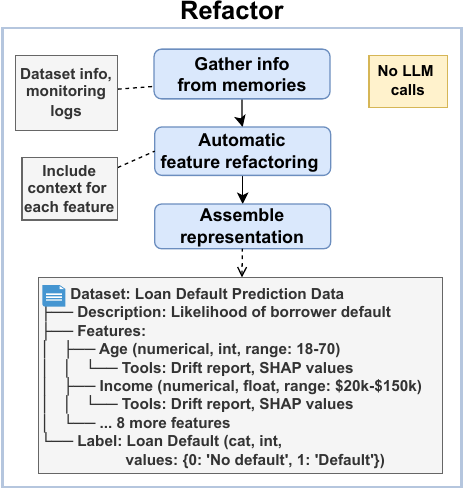}
\caption{Refactor step gathers and structures information from memories without LLM calls.}
\label{fig:refactor}
\end{figure}

\textbf{Refactor} improves data representation by gathering information from memories, performing automatic feature refactoring without LLM calls, and assembling a structured representation with context for each feature. The output is a well-organized dataset representation, as shown in Figure \ref{fig:refactor}.

\begin{figure}[h]
\centering
\includegraphics[width=0.7\linewidth]{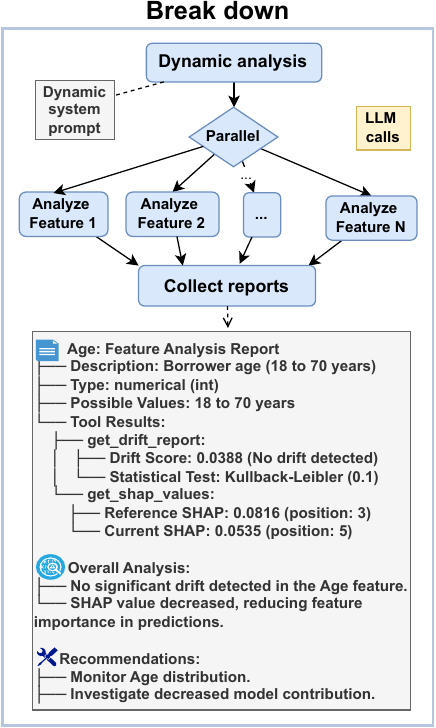}
\caption{Break Down step analyzes features in parallel using LLM calls.}
\label{fig:break_down}
\end{figure}

\textbf{Break Down} decomposes complex information for detailed analysis. As illustrated in Figure \ref{fig:break_down}, this step uses a dynamic system prompt to guide parallel feature analysis, collecting detailed insights and integrating tool results like drift reports and SHAP \cite{lundberg2017unifiedapproachinterpretingmodel} values.

\begin{figure}[h]
\centering
\includegraphics[width=0.45\linewidth]{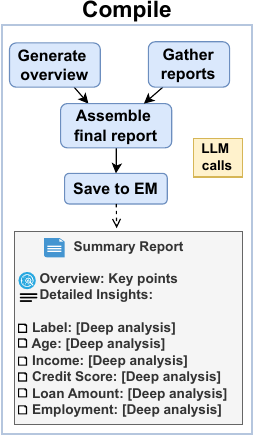}
\caption{Compile step generates the final comprehensive report.}
\label{fig:compile}
\end{figure}

\textbf{Compile} integrates all insights into clear, interpretable outputs. As shown in Figure \ref{fig:compile}, it generates an overview, gathers feature reports, assembles a comprehensive report using LLM calls, and saves it to Episodic Memory.

The resulting report provides a holistic view through summary, key points, and detailed feature insights. By leveraging structured memory, feature engineering principles, and adaptive learning, we enhance the LLM's ability to provide accurate, context-aware insights. The LLM serves as a reasoning engine enabling multi-step analysis, and our LLM-agnostic approach allows flexibility in model selection, with even small models performing well at certain tasks.

\subsection{Operational Characteristics}
CAMA is designed to be flexible in its deployment and operation. While primarily user-triggered, it can be integrated into automated workflows through cron jobs or DevOps pipelines. The architecture supports composability, allowing multiple CAMA agents to be chained together in a multi-agent pipeline. For example, when one agent detects significant drift, it can trigger another agent to analyze the drift's impact and suggest model improvements. This multi-agent capability enables sophisticated monitoring workflows where each agent specializes in different aspects of the monitoring process.

\section{Experimental Setup}
\subsection{Models and Datasets}
\subsubsection{LLM Models}
We evaluated four LLMs to compare their reasoning capabilities for interpreting monitoring tool outputs:
\begin{itemize}
\item \textbf{Llama-3.2-1b}: A 1 billion parameter model, representing smaller, more resource-efficient LLMs.
\item \textbf{Llama3-8b}: An 8 billion parameter model with an 8,192 token context window, representing medium-sized LLMs.
\item \textbf{Llama3-70b}: A 70 billion parameter model with an 8,192 token context window, representing large-scale LLMs.
\item \textbf{GPT-4o-mini}: A multimodal model with a larger context window, representing state-of-the-art performance. The exact model size has not been published.
\end{itemize}
All models were used with a temperature setting of 0 to ensure consistency in responses. This range of models allows us to evaluate the performance of our approach across different model sizes and architectures, from resource-efficient options to state-of-the-art performers.
The inclusion of Llama-3.2-1b enables us to assess the effectiveness of our approach on smaller models, which are often preferred in resource-constrained environments. The comparison across these models provides insights into how our cognitive architecture scales with different LLM capabilities and sizes.

\subsubsection{Datasets}
We curated three datasets representing diverse domains and drift scenarios, each with varying levels of complexity. These datasets were synthetically generated using GPT-4o and subsequently refined to ensure realism and consistency. The refinement process addressed potential issues such as nonsensical drifts, ensured correct value ranges (e.g., for age), and maintained appropriate type interdependencies and distributions. Detailed information about the dataset generation and refinement process is available in the supplementary file. The datasets are as follows:
\noindent\textbf{Loan Default Prediction (Easy)}: A financial dataset with 10 features (7 numerical, 3 categorical). This dataset is considered easy due to its straightforward feature set and clear relationships between variables such as income, credit score, and loan default probability. Features include Age (18-70), Income (\$20K-\$150K), Credit Score (300-850), Loan Amount (\$1K-\$50K), and categorical variables like Home Ownership (Rent/Own/Mortgage).

\noindent\textbf{Eligibility Simulation (Medium)}: A dataset with 5 features (2 numerical, 3 categorical). It is classified as medium difficulty due to the interplay between socioeconomic factors and eligibility criteria, requiring more nuanced interpretation. The numerical features are designed with realistic ranges based on socioeconomic indicators.

\noindent\textbf{Chronic Condition Prediction (Difficult)}: A healthcare dataset with 10 features (6 numerical, 4 categorical). This dataset is considered difficult due to the complex interactions between various health indicators, lifestyle factors, and socioeconomic variables in predicting chronic conditions. Numerical features include health metrics and lifestyle indicators with clinically relevant ranges.

Each dataset was designed to exhibit realistic distribution drifts, challenging the monitoring capabilities of our system and baselines across different complexity levels. It is important to note that while the original datasets contain 1,000 samples each, our monitoring tools utilize only 100 samples for drift calculation and feature attribution. This approach ensures that the sample size does not significantly impact the monitoring process, allowing for efficient analysis regardless of the original dataset size.

\subsection{Evaluation Methodology}
Our evaluation follows the MMLU (Measuring Massive Multitask Language Understanding) approach \cite{hendrycks2021measuringmassivemultitasklanguage}, using four key metrics and a structured evaluation process:


\begin{enumerate}
\item\textbf{Accuracy (↑)}: Percentage of correct answers to multi-choice questions, measuring report quality.

\item\textbf{Unknown Ratio (↓)}: Percentage of "I DON'T KNOW" responses, indicating information gaps.

\item\textbf{Tokens (↓)}: Number of tokens generated, measuring conciseness and efficiency.

\item\textbf{Time (↓)}: Elapsed time for report generation, affected by model latency.
\end{enumerate}
All metrics are reported with mean and standard deviation across multiple runs.

The evaluation consists of five steps:
\begin{enumerate}
\item \textbf{Ground Truth Creation}: Comprehensive reports using raw data and monitoring tools, including executive summaries, dataset synopses, and detailed analyses.

\item \textbf{Question Generation}: 39 multi-choice questions per dataset created by GPT-4o, covering various aspects of the reports.

\item \textbf{Report Generation}: Reports generated using CAMA and baseline methods across all LLM models.

\item \textbf{Evaluation}: GPT-4o as impartial judge answers pre-defined questions for each report.

\item \textbf{Metric Calculation}: Comparison of GPT-4o's answers against ground truth for accuracy and unknown ratio, plus token and time measurements.
\end{enumerate}

This systematic approach ensures fair comparison across methods and models, with robust and consistent performance assessment.

\subsection{Comparison Methods}
We compared CAMA against six alternative methods:

\begin{itemize}
\item Standard (I/O): Direct input-output prompting.
\item Chain of Thought (CoT): Prompting with intermediate reasoning steps.
\item Reflection: Iterative self-correction approach.
\item ReAct: Reasoning and acting framework.
\item Self-Discover: Autonomous reasoning structure generation.
\item Plan \& Execute: Structured planning and execution approach.
\end{itemize}

\subsection{Implementation Details}
Our implementation executed LLM models across two distinct experimental setups:
\begin{itemize}
\item \textbf{Large Language Models API}: We integrated GROQ\footnote{https://groq.com/}, OpenRouter\footnote{https://openrouter.ai/}, and OpenAI APIs to access large-scale language model inference capabilities.
\item \textbf{Small Language Models API}: To access inference capabilities of CPU-optimized smaller language models, we leveraged the Intel® Optimized Stack on a Lenovo cluster with dual Intel® Xeon® Platinum 8360Y processors @ 2.40GHz and 256 GB RAM, running Ubuntu 22.04 LTS. This setup utilized the Intel® AI Analytics Toolkit (via Docker image intel/oneapi-aikit:devel-ubuntu22.04\footnote{https://hub.docker.com/r/intel/oneapi-aikit}) and Intel® Extension for PyTorch\footnote{https://github.com/intel/intel-extension-for-pytorch} 1.12.100+cpu.
\end{itemize}
We also leveraged other key technologies and libraries:
\begin{itemize}
\item \textbf{Cognitive Architecture}: Implemented using Langchain\footnote{https://www.langchain.com/} and Langgraph\footnote{https://langchain-ai.github.io/langgraph/} for flexible and efficient workflow management.
\item \textbf{Data Representation}: Docarray\footnote{https://docs.docarray.org/} was employed for handling unstructured data representations.
\item \textbf{Monitoring Tools}: We utilized Alibi Detect\footnote{https://github.com/SeldonIO/alibi-detect} for calculating drift scores and feature attributions.
\end{itemize}
All methods were implemented using Python 3.9, ensuring consistency in environment and dependencies for fair comparisons

\subsection{Reproducibility}
To facilitate reproducibility, we have released our codebase\footnote{\url{https://github.com/gusseppe/cognitive_architecture_checker/}}, including dataset generation scripts, model implementations, evaluation pipelines, and configuration files for all experiments.

\section{Results and Discussion}
\begin{figure}[t]
\centering
\includegraphics[width=0.8\columnwidth]{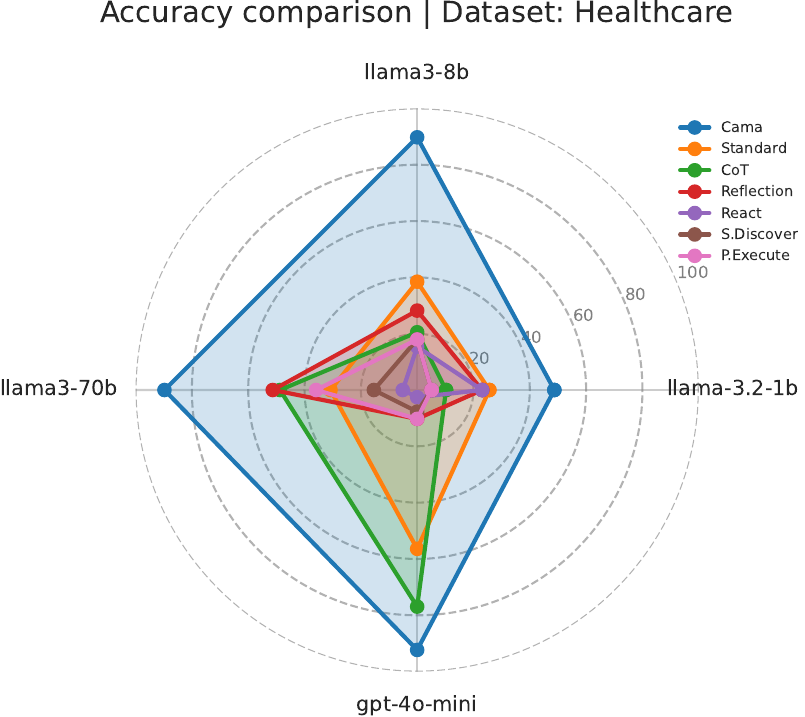}
\caption{Accuracy comparison for the Healthcare dataset across different models and methods.}
\label{fig:radar_healthcare}
\end{figure}

\begin{figure}[t]
\centering
\includegraphics[width=0.8\columnwidth]{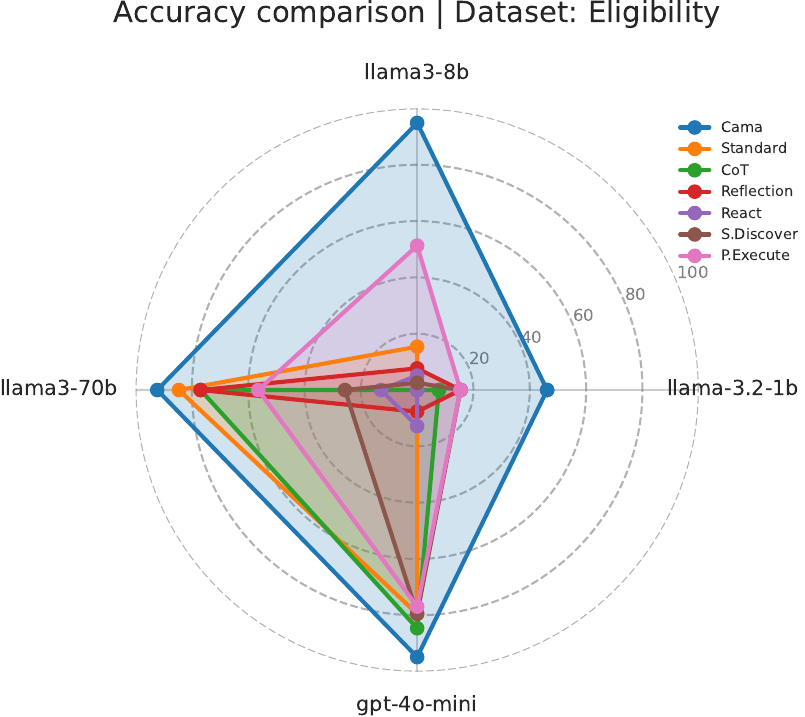}
\caption{Accuracy comparison for the Eligibility dataset across different models and methods.}
\label{fig:radar_eligibility}
\end{figure}

\begin{figure}[t]
\centering
\includegraphics[width=0.8\columnwidth]{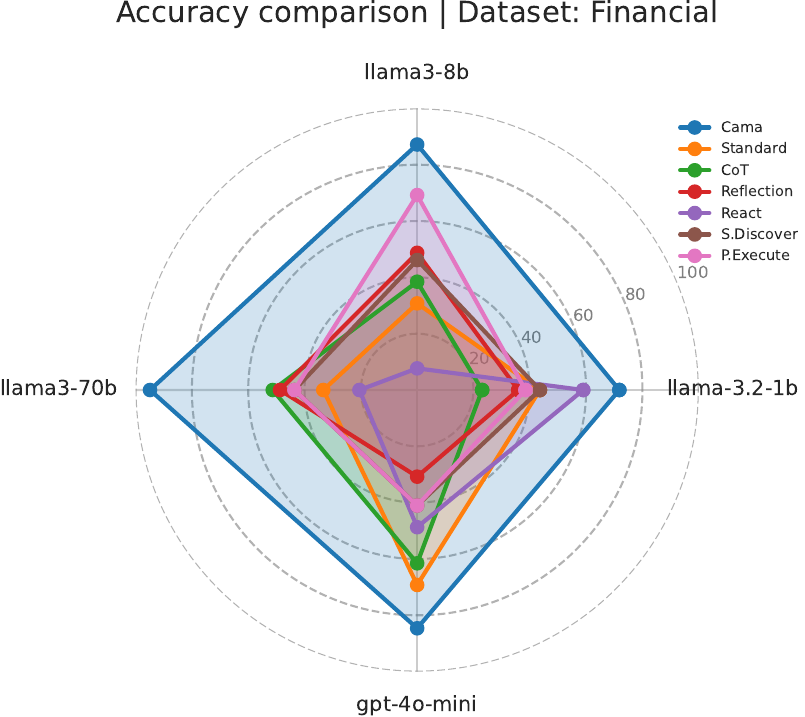}
\caption{Accuracy comparison for the Financial dataset across different models and methods.}
\label{fig:radar_financial}
\end{figure}

Our proposed method, CAMA, demonstrates superior performance across all evaluated scenarios, as shown in Table \ref{tab:performance_comparison} and Figures \ref{fig:radar_healthcare}, \ref{fig:radar_eligibility}, and \ref{fig:radar_financial}. These results provide a comprehensive comparison of our approach against other established methods across different datasets and model sizes.

\begin{table*}[htbp]
\centering
\caption{Average Performance Comparison Across All Datasets}
\label{tab:performance_comparison}
\resizebox{\textwidth}{!}{%
\begin{tabular}{llccccccc}
\toprule
\textbf{Model} & \textbf{Metric} & \textbf{CAMA} & \textbf{Standard} & \textbf{CoT} & \textbf{Reflection} & \textbf{React} & \textbf{S.Discover} & \textbf{P.Execute} \\
\midrule
\multirow{4}{*}{\textbf{llama-3.2-1b}} & \textbf{Accuracy (↑)} & \textbf{55.6 ± 8.1} & 23.1 ± 12.6 & 13.7 ± 4.8 & 24.8 ± 6.0 & 27.4 ± 17.2 & 21.4 ± 11.5 & 19.7 ± 9.9 \\
 & \textbf{Unknown (↓)} & \textbf{35.9 ± 5.4} & 76.9 ± 12.6 & 85.5 ± 4.5 & 70.1 ± 8.9 & 69.2 ± 19.4 & 77.8 ± 12.3 & 80.3 ± 9.9 \\
 & \textbf{Tokens (↓)} & 24.9 ± 5.2 & 4.9 ± 0.2 & \textbf{3.8 ± 0.8} & 14.2 ± 0.5 & 23.6 ± 9.6 & 17.3 ± 0.7 & 23.7 ± 7.0 \\
 & \textbf{Time (↓)} & 5.6 ± 0.3 & \textbf{1.5 ± 0.6} & 2.0 ± 0.4 & 19.7 ± 16.2 & 35.2 ± 23.5 & 61.0 ± 27.7 & 46.6 ± 2.2 \\
\midrule
\multirow{4}{*}{\textbf{llama3-8b}} & \textbf{Accuracy (↑)} & \textbf{90.6 ± 2.3} & 28.2 ± 6.8 & 19.7 ± 11.1 & 28.2 ± 11.8 & 9.4 ± 3.1 & 22.2 ± 12.8 & 46.1 ± 15.0 \\
 & \textbf{Unknown (↓)} & \textbf{0.9 ± 0.9} & 70.1 ± 8.2 & 80.3 ± 11.1 & 68.4 ± 12.8 & 89.8 ± 3.9 & 76.1 ± 13.4 & 47.9 ± 16.6 \\
 & \textbf{Tokens (↓)} & 19.8 ± 4.1 & 5.0 ± 0.1 & \textbf{4.8 ± 0.2} & 10.0 ± 3.3 & 5.7 ± 1.3 & 20.2 ± 3.2 & 34.9 ± 3.7 \\
 & \textbf{Time (↓)} & 5.2 ± 0.3 & \textbf{1.1 ± 0.1} & 3.6 ± 2.6 & 11.7 ± 7.8 & 6.6 ± 4.2 & 22.7 ± 11.1 & 38.3 ± 12.0 \\
\midrule
\multirow{4}{*}{\textbf{llama3-70b}} & \textbf{Accuracy (↑)} & \textbf{92.3 ± 1.5} & 49.6 ± 17.5 & 59.0 ± 9.0 & 59.0 ± 9.0 & 12.8 ± 4.4 & 28.2 ± 8.2 & 45.3 ± 6.0 \\
 & \textbf{Unknown (↓)} & \textbf{0.0 ± 0.0} & 42.7 ± 21.4 & 35.9 ± 12.8 & 32.5 ± 12.6 & 85.5 ± 3.1 & 66.7 ± 8.3 & 50.4 ± 7.6 \\
 & \textbf{Tokens (↓)} & 19.1 ± 2.2 & \textbf{4.0 ± 0.9} & 4.3 ± 1.0 & 12.3 ± 1.7 & 7.1 ± 1.5 & 19.5 ± 3.4 & 37.3 ± 25.8 \\
 & \textbf{Time (↓)} & 72.2 ± 11.7 & \textbf{12.5 ± 5.2} & 35.3 ± 6.5 & 122.2 ± 13.3 & 21.2 ± 2.2 & 123.6 ± 29.9 & 249.0 ± 128.9 \\
\midrule
\multirow{4}{*}{\textbf{gpt-4o-mini}} & \textbf{Accuracy (↑)} & \textbf{90.6 ± 3.1} & 68.4 ± 6.7 & 74.3 ± 6.8 & 16.3 ± 7.3 & 21.4 ± 14.0 & 42.7 ± 20.7 & 42.7 ± 19.2 \\
 & \textbf{Unknown (↓)} & \textbf{0.0 ± 0.0} & 29.1 ± 6.0 & 19.7 ± 9.5 & 82.9 ± 8.1 & 75.2 ± 15.2 & 56.4 ± 20.0 & 51.3 ± 16.0 \\
 & \textbf{Tokens (↓)} & 20.2 ± 4.0 & 4.8 ± 0.8 & \textbf{4.5 ± 1.1} & 13.6 ± 2.3 & 7.8 ± 1.7 & 22.6 ± 1.8 & 495.9 ± 10.8 \\
 & \textbf{Time (↓)} & 27.2 ± 5.5 & \textbf{19.1 ± 3.6} & 23.9 ± 2.9 & 57.0 ± 2.9 & 25.9 ± 3.4 & 101.1 ± 37.0 & 340.3 ± 64.1 \\
\bottomrule
\end{tabular}
}
\end{table*}

\subsection{Performance Analysis}
CAMA consistently outperforms all other methods across different model sizes and datasets:
\begin{itemize}
\item \textbf{Accuracy:} CAMA achieves the highest accuracy for all models, with particularly impressive results for larger models. For llama3-70b, CAMA reaches 92.3\% ± 1.5\% accuracy, significantly outperforming the next best method (CoT and Reflection, both at 59.0\% ± 9.0\%).
\item \textbf{Unknown Ratio:} Our method maintains the lowest unknown ratios, reaching 0.0\% for both llama3-70b and gpt-4o-mini, indicating comprehensive and informative reports.
\item \textbf{Tokens and Time:} While CAMA generally generates more tokens compared to simpler methods, the trade-off in performance is substantial. The increased token count is justified by the significant gains in accuracy and reduction in unknown responses.
\end{itemize}
The radar plots (Figures \ref{fig:radar_healthcare}, \ref{fig:radar_eligibility}, and \ref{fig:radar_financial}) demonstrate CAMA's superior performance across all datasets and model sizes. In each plot, CAMA's accuracy (represented by the blue area) consistently extends further from the center than other methods, indicating higher accuracy across all scenarios.

\subsection{Comparative Analysis}
CAMA's performance contrasts sharply with other approaches:
\begin{itemize}
\item \textbf{Standard (I/O):} Shows moderate accuracy but high unknown ratios, particularly for smaller models.
\item \textbf{Chain of Thought (CoT):} Performs well with larger models but struggles with smaller ones, suggesting its effectiveness depends on model size.
\item \textbf{Reflection:} Shows inconsistent performance across model sizes and datasets.
\item \textbf{ReAct:} Performs poorly across all model sizes, indicating potential issues with its reasoning and action framework in this context.
\item \textbf{Self Discover:} Demonstrates low accuracy and high unknown ratios, suggesting difficulties in autonomous reasoning structure generation for this task.
\item \textbf{Plan \& Execute:} Shows moderate performance with some models but struggles with consistency across datasets and model sizes.
\end{itemize}
These results reveal that while some methods (like CoT) work better with larger models, CAMA consistently outperforms all methods across model sizes and datasets. This suggests that our cognitive architecture effectively leverages the strengths of various models while providing additional structure and guidance.

\subsection{Model Size Impact}
The impact of model size on performance is evident:

\begin{itemize}
\item \textbf{llama-3.2-1b:} Even with this smaller model, CAMA achieves 55.6\% ± 8.1\% accuracy, significantly outperforming other methods.
\item \textbf{llama3-8b:} CAMA's performance jumps to 90.6\% ± 2.3\% accuracy, demonstrating substantial improvement with increased model size.
\item \textbf{llama3-70b and gpt-4o-mini:} These larger models show CAMA's peak performance, with accuracies of 92.3\% ± 1.5\% and 90.6\% ± 3.1\% respectively.
\end{itemize}

This trend suggests that while CAMA benefits from larger models, it can still provide significant improvements even with smaller, more resource-efficient models.

\subsection{Dataset-specific Performance}
Figures \ref{fig:radar_healthcare}, \ref{fig:radar_eligibility}, and \ref{fig:radar_financial} illustrate CAMA's consistent superior performance across different datasets:
\begin{itemize}
\item \textbf{Healthcare:} CAMA shows a clear advantage, particularly with llama3-8b and llama3-70b models.
\item \textbf{Eligibility:} The performance gap is most pronounced in this dataset, with CAMA significantly outperforming other methods across all model sizes.
\item \textbf{Financial:} While the performance gap narrows for some models (particularly gpt-4o-mini), CAMA still maintains a clear lead.
\end{itemize}
These dataset-specific results demonstrate CAMA's versatility and robustness across different domains and data complexities.

\subsection{Ablation Study}
\label{subsec:ablation}
To evaluate the importance of each component in our proposed architecture, we conducted an ablation study using the financial dataset and the llama3-8b model. Table \ref{tab:ablation} presents the results.

\begin{table}[t]
\centering
\caption{Ablation Study on financial dataset using llama3-8b}
\label{tab:ablation}
\begin{tabular}{lc}
\toprule
\textbf{Configuration} & \textbf{Accuracy (\%)} \\
\midrule
Full Pipeline & 88.30 \\
\midrule
w/o Refactor & 20.51 \\
w/o Break Down & 23.08 \\
w/o Compile & 69.23\\
\bottomrule
\end{tabular}
\end{table}

The ablation study reveals several crucial insights:

\begin{itemize}
\item \textbf{Component Interdependence:} Removing any single component results in a significant performance decrease, with accuracy dropping by at least 19 percentage points (in the case of removing Compile). This demonstrates the interdependence of all components in our pipeline.
\item \textbf{Critical Components:} The Refactor and Break Down components appear most critical, with their removal causing accuracy drops more than 67.8 and 65.2 percentage points, respectively. Refactor's importance stems from its context-aware filtering that provides essential feature representations, while Break Down enables focused analysis of individual features, reducing cognitive load on LLMs. This aligns with our design philosophy of improving data representation and detailed analysis before synthesis.

\item \textbf{Compile Importance:} While less impactful than Refactor and Break Down, the Compile step still contributes significantly to the overall performance, improving accuracy by 19.07 percentage points.

\item \textbf{Pipeline Robustness:} The substantial accuracy drop when removing any component indicates that each step is crucial for optimal performance. This justifies the complexity of our full pipeline and highlights the importance of maintaining all components for best results.
\end{itemize}

\subsection{Discussion}
The superior performance of CAMA across different model sizes and metrics underscores the effectiveness of our adaptive cognitive architecture. Several factors contribute to these outcomes:
\begin{enumerate}
\item \textbf{Effective use of memory modules:} Our method's structured use of Procedural, Episodic, Semantic, and Working Memory ensures efficient retention and utilization of relevant information throughout the analysis process.
\item \textbf{Feature engineering-inspired approach:} The Refactor, Break Down, and Compile steps allow our method to handle complex scenarios and evolving data effectively, as evidenced by the significant performance drops when any of these components are removed.

\item \textbf{Adaptability across model sizes:} CAMA's consistent performance improvements across different model sizes (from llama-3.2-1b to llama3-70b) demonstrate its ability to effectively leverage the strengths of various LLMs while providing additional guidance and structure.
\item \textbf{Performance-token trade-off:} While CAMA's approach consumes more tokens due to feature-specific context preservation and independent reasoning chains, this design choice is justified by the significant performance gains across all metrics.

\end{enumerate}
The ablation study further confirms the importance of each component in our pipeline, demonstrating that the full architecture is necessary to achieve optimal performance.

\section{Conclusion}

We presented CAMA, an adaptive cognitive architecture for monitoring ML models using LLMs. Leveraging multiple memory types, our method automates monitoring and reporting, providing accurate and actionable insights. Experimental results demonstrate CAMA's superior performance across various metrics, model sizes, and datasets. With llama3-70b, CAMA achieved 92.3\% ± 1.5\% accuracy, outperforming the next best method by 33.3 percentage points. For llama3-8b, CAMA's accuracy (90.6\% ± 2.3\%) surpassed the next best method by 44.5 percentage points. Even with llama-3.2-1b, CAMA showed robust performance (55.6\% ± 8.1\% accuracy).
The three-step process of Refactor, Break Down, and Compile enables efficient evaluations, adapting to diverse data drift scenarios. Our ablation study highlights the necessity of each component in our pipeline. CAMA's consistent superior performance across Healthcare, Eligibility, and Financial datasets demonstrates its versatility and adaptability. This, combined with its effectiveness across various model sizes, positions CAMA as a powerful tool for enhancing ML model monitoring in production environments, contributing to more reliable and transparent AI systems.

\section{Limitations and Future Work}
While CAMA demonstrates significant improvements in ML model monitoring, our current study has some limitations. The evaluation was conducted on a specific set of datasets and LLM architectures, which may limit the generalizability of our findings. Additionally, the computational requirements and processing times of our method in comparison to simpler approaches warrant further investigation.
Future research should focus on expanding the evaluation of CAMA across a broader range of LLM architectures, monitoring scenarios, and domains to establish wider applicability. Investigating techniques to optimize the implementation for reduced computational overhead and processing time could enhance CAMA's efficiency. These advancements would contribute to making CAMA more robust and adaptable for diverse real-world ML monitoring scenarios, ultimately leading to more reliable ML systems.

\begin{acks}
We thank Lenovo for providing the technical infrastructure to run the experiments in this paper. This work was partially supported by Lenovo and Intel® as part of the Lenovo AI Innovators University Research program, by Spanish Ministry of Science (MICINN), the Research State Agency (AEI) and European Regional Development Funds (ERDF/FEDER) under contracts PID2019-107255GB-C22 and PID2021-126248OB-I00, and by the Generalitat de Catalunya (AGAUR) under contract 2021-SGR-00478.
\end{acks}



\balance

\bibliographystyle{ACM-Reference-Format} 

\bibliography{sample}


\begin{thebibliography}{17}


\ifx \showCODEN    \undefined \def \showCODEN     #1{\unskip}     \fi
\ifx \showDOI      \undefined \def \showDOI       #1{#1}\fi
\ifx \showISBNx    \undefined \def \showISBNx     #1{\unskip}     \fi
\ifx \showISBNxiii \undefined \def \showISBNxiii  #1{\unskip}     \fi
\ifx \showISSN     \undefined \def \showISSN      #1{\unskip}     \fi
\ifx \showLCCN     \undefined \def \showLCCN      #1{\unskip}     \fi
\ifx \shownote     \undefined \def \shownote      #1{#1}          \fi
\ifx \showarticletitle \undefined \def \showarticletitle #1{#1}   \fi
\ifx \showURL      \undefined \def \showURL       {\relax}        \fi
\providecommand\bibfield[2]{#2}
\providecommand\bibinfo[2]{#2}
\providecommand\natexlab[1]{#1}
\providecommand\showeprint[2][]{arXiv:#2}

\bibitem[\protect\citeauthoryear{Brown, Mann, Ryder, Subbiah, Kaplan, Dhariwal, Neelakantan, Shyam, Sastry, Askell, Agarwal, Herbert-Voss, Krueger, Henighan, Child, Ramesh, Ziegler, Wu, Winter, Hesse, Chen, Sigler, Litwin, Gray, Chess, Clark, Berner, McCandlish, Radford, Sutskever, and Amodei}{Brown et~al\mbox{.}}{2020}]%
        {brown2020language}
\bibfield{author}{\bibinfo{person}{Tom Brown}, \bibinfo{person}{Benjamin Mann}, \bibinfo{person}{Nick Ryder}, \bibinfo{person}{Melanie Subbiah}, \bibinfo{person}{Jared~D Kaplan}, \bibinfo{person}{Prafulla Dhariwal}, \bibinfo{person}{Arvind Neelakantan}, \bibinfo{person}{Pranav Shyam}, \bibinfo{person}{Girish Sastry}, \bibinfo{person}{Amanda Askell}, \bibinfo{person}{Sandhini Agarwal}, \bibinfo{person}{Ariel Herbert-Voss}, \bibinfo{person}{Gretchen Krueger}, \bibinfo{person}{Tom Henighan}, \bibinfo{person}{Rewon Child}, \bibinfo{person}{Aditya Ramesh}, \bibinfo{person}{Daniel Ziegler}, \bibinfo{person}{Jeffrey Wu}, \bibinfo{person}{Clemens Winter}, \bibinfo{person}{Chris Hesse}, \bibinfo{person}{Mark Chen}, \bibinfo{person}{Eric Sigler}, \bibinfo{person}{Mateusz Litwin}, \bibinfo{person}{Scott Gray}, \bibinfo{person}{Benjamin Chess}, \bibinfo{person}{Jack Clark}, \bibinfo{person}{Christopher Berner}, \bibinfo{person}{Sam McCandlish}, \bibinfo{person}{Alec Radford}, \bibinfo{person}{Ilya Sutskever}, {and}
  \bibinfo{person}{Dario Amodei}.} \bibinfo{year}{2020}\natexlab{}.
\newblock \showarticletitle{Language Models are Few-Shot Learners}. In \bibinfo{booktitle}{\emph{Advances in Neural Information Processing Systems}}, \bibfield{editor}{\bibinfo{person}{H.~Larochelle}, \bibinfo{person}{M.~Ranzato}, \bibinfo{person}{R.~Hadsell}, \bibinfo{person}{M.F. Balcan}, {and} \bibinfo{person}{H.~Lin}} (Eds.), Vol.~\bibinfo{volume}{33}. \bibinfo{publisher}{Curran Associates, Inc.}, \bibinfo{pages}{1877--1901}.
\newblock


\bibitem[\protect\citeauthoryear{Cosentino, Belyaeva, Liu, Furlotte, Yang, Lee, Schenck, Patel, Cui, Schneider, Bryant, Gomes, Jiang, Lee, Liu, Perez, Rogers, Speed, Tailor, Walker, Yu, Althoff, Heneghan, Hernandez, Malhotra, Stern, Matias, Corrado, Patel, Shetty, Zhan, Prabhakara, McDuff, and McLean}{Cosentino et~al\mbox{.}}{2024}]%
        {cosentino2024personal}
\bibfield{author}{\bibinfo{person}{Justin Cosentino}, \bibinfo{person}{Anastasiya Belyaeva}, \bibinfo{person}{Xin Liu}, \bibinfo{person}{Nicholas~A. Furlotte}, \bibinfo{person}{Zhun Yang}, \bibinfo{person}{Chace Lee}, \bibinfo{person}{Erik Schenck}, \bibinfo{person}{Yojan Patel}, \bibinfo{person}{Jian Cui}, \bibinfo{person}{Logan~Douglas Schneider}, \bibinfo{person}{Robby Bryant}, \bibinfo{person}{Ryan~G. Gomes}, \bibinfo{person}{Allen Jiang}, \bibinfo{person}{Roy Lee}, \bibinfo{person}{Yun Liu}, \bibinfo{person}{Javier Perez}, \bibinfo{person}{Jameson~K. Rogers}, \bibinfo{person}{Cathy Speed}, \bibinfo{person}{Shyam Tailor}, \bibinfo{person}{Megan Walker}, \bibinfo{person}{Jeffrey Yu}, \bibinfo{person}{Tim Althoff}, \bibinfo{person}{Conor Heneghan}, \bibinfo{person}{John Hernandez}, \bibinfo{person}{Mark Malhotra}, \bibinfo{person}{Leor Stern}, \bibinfo{person}{Yossi Matias}, \bibinfo{person}{Greg~S. Corrado}, \bibinfo{person}{Shwetak Patel}, \bibinfo{person}{Shravya Shetty}, \bibinfo{person}{Jiening Zhan},
  \bibinfo{person}{Shruthi Prabhakara}, \bibinfo{person}{Daniel McDuff}, {and} \bibinfo{person}{Cory~Y. McLean}.} \bibinfo{year}{2024}\natexlab{}.
\newblock \bibinfo{title}{Towards a Personal Health Large Language Model}.
\newblock
\newblock
\showeprint[arxiv]{2406.06474}~[cs.AI]
\urldef\tempurl%
\url{https://arxiv.org/abs/2406.06474}
\showURL{%
\tempurl}


\bibitem[\protect\citeauthoryear{Eck, Kabakci-Zorlu, Chen, Savard, and Bao}{Eck et~al\mbox{.}}{2022}]%
        {eck2022monitoring}
\bibfield{author}{\bibinfo{person}{Bradley Eck}, \bibinfo{person}{Duygu Kabakci-Zorlu}, \bibinfo{person}{Yan Chen}, \bibinfo{person}{France Savard}, {and} \bibinfo{person}{Xiaowei Bao}.} \bibinfo{year}{2022}\natexlab{}.
\newblock \showarticletitle{A monitoring framework for deployed machine learning models with supply chain examples}. In \bibinfo{booktitle}{\emph{2022 IEEE International Conference on Big Data (Big Data)}}. \bibinfo{pages}{2231--2238}.
\newblock
\urldef\tempurl%
\url{https://doi.org/10.1109/BigData55660.2022.10020394}
\showDOI{\tempurl}


\bibitem[\protect\citeauthoryear{Heinrichs}{Heinrichs}{2023}]%
        {monitoring_ml_models}
\bibfield{author}{\bibinfo{person}{Florian Heinrichs}.} \bibinfo{year}{2023}\natexlab{}.
\newblock \bibinfo{title}{Monitoring Machine Learning Models: Online Detection of Relevant Deviations}.
\newblock
\newblock
\showeprint[arxiv]{2309.15187}~[cs.LG]
\urldef\tempurl%
\url{https://arxiv.org/abs/2309.15187}
\showURL{%
\tempurl}


\bibitem[\protect\citeauthoryear{Hendrycks, Burns, Basart, Zou, Mazeika, Song, and Steinhardt}{Hendrycks et~al\mbox{.}}{2021}]%
        {hendrycks2021measuringmassivemultitasklanguage}
\bibfield{author}{\bibinfo{person}{Dan Hendrycks}, \bibinfo{person}{Collin Burns}, \bibinfo{person}{Steven Basart}, \bibinfo{person}{Andy Zou}, \bibinfo{person}{Mantas Mazeika}, \bibinfo{person}{Dawn Song}, {and} \bibinfo{person}{Jacob Steinhardt}.} \bibinfo{year}{2021}\natexlab{}.
\newblock \bibinfo{title}{Measuring Massive Multitask Language Understanding}.
\newblock
\newblock
\showeprint[arxiv]{2009.03300}~[cs.CY]
\urldef\tempurl%
\url{https://arxiv.org/abs/2009.03300}
\showURL{%
\tempurl}


\bibitem[\protect\citeauthoryear{Kojima, Gu, Reid, Matsuo, and Iwasawa}{Kojima et~al\mbox{.}}{2024}]%
        {kojima2022large}
\bibfield{author}{\bibinfo{person}{Takeshi Kojima}, \bibinfo{person}{Shixiang~Shane Gu}, \bibinfo{person}{Machel Reid}, \bibinfo{person}{Yutaka Matsuo}, {and} \bibinfo{person}{Yusuke Iwasawa}.} \bibinfo{year}{2024}\natexlab{}.
\newblock \showarticletitle{Large language models are zero-shot reasoners}. In \bibinfo{booktitle}{\emph{Proceedings of the 36th International Conference on Neural Information Processing Systems}} (New Orleans, LA, USA) \emph{(\bibinfo{series}{NIPS'22})}. \bibinfo{publisher}{Curran Associates Inc.}, \bibinfo{address}{Red Hook, NY, USA}, Article \bibinfo{articleno}{1613}, \bibinfo{numpages}{15}~pages.
\newblock
\showISBNx{9781713871088}


\bibitem[\protect\citeauthoryear{Lipton, Wang, and Smola}{Lipton et~al\mbox{.}}{2018}]%
        {lipton2018detecting}
\bibfield{author}{\bibinfo{person}{Zachary~C. Lipton}, \bibinfo{person}{Yu-Xiang Wang}, {and} \bibinfo{person}{Alex Smola}.} \bibinfo{year}{2018}\natexlab{}.
\newblock \bibinfo{title}{Detecting and Correcting for Label Shift with Black Box Predictors}.
\newblock
\newblock
\showeprint[arxiv]{1802.03916}~[cs.LG]
\urldef\tempurl%
\url{https://arxiv.org/abs/1802.03916}
\showURL{%
\tempurl}


\bibitem[\protect\citeauthoryear{Lundberg and Lee}{Lundberg and Lee}{2017}]%
        {lundberg2017unifiedapproachinterpretingmodel}
\bibfield{author}{\bibinfo{person}{Scott~M. Lundberg} {and} \bibinfo{person}{Su-In Lee}.} \bibinfo{year}{2017}\natexlab{}.
\newblock \showarticletitle{A unified approach to interpreting model predictions}. In \bibinfo{booktitle}{\emph{Proceedings of the 31st International Conference on Neural Information Processing Systems}} (Long Beach, California, USA) \emph{(\bibinfo{series}{NIPS'17})}. \bibinfo{publisher}{Curran Associates Inc.}, \bibinfo{address}{Red Hook, NY, USA}, \bibinfo{pages}{4768–4777}.
\newblock
\showISBNx{9781510860964}


\bibitem[\protect\citeauthoryear{Rabanser, G\"{u}nnemann, and Lipton}{Rabanser et~al\mbox{.}}{2019}]%
        {rabanser2019failing}
\bibfield{author}{\bibinfo{person}{Stephan Rabanser}, \bibinfo{person}{Stephan G\"{u}nnemann}, {and} \bibinfo{person}{Zachary~C. Lipton}.} \bibinfo{year}{2019}\natexlab{}.
\newblock \showarticletitle{Failing loudly: an empirical study of methods for detecting dataset shift}. In \bibinfo{booktitle}{\emph{Proceedings of the 33rd International Conference on Neural Information Processing Systems}}. \bibinfo{publisher}{Curran Associates Inc.}, \bibinfo{address}{Red Hook, NY, USA}, Article \bibinfo{articleno}{125}, \bibinfo{numpages}{13}~pages.
\newblock


\bibitem[\protect\citeauthoryear{Shinn, Cassano, Gopinath, Narasimhan, and Yao}{Shinn et~al\mbox{.}}{2024}]%
        {shinn2023reflexion}
\bibfield{author}{\bibinfo{person}{Noah Shinn}, \bibinfo{person}{Federico Cassano}, \bibinfo{person}{Ashwin Gopinath}, \bibinfo{person}{Karthik Narasimhan}, {and} \bibinfo{person}{Shunyu Yao}.} \bibinfo{year}{2024}\natexlab{}.
\newblock \showarticletitle{Reflexion: language agents with verbal reinforcement learning}. In \bibinfo{booktitle}{\emph{Proceedings of the 37th International Conference on Neural Information Processing Systems}} (New Orleans, LA, USA) \emph{(\bibinfo{series}{NIPS'23})}. \bibinfo{publisher}{Curran Associates Inc.}, \bibinfo{address}{Red Hook, NY, USA}, Article \bibinfo{articleno}{377}, \bibinfo{numpages}{19}~pages.
\newblock


\bibitem[\protect\citeauthoryear{Sumers, Yao, Narasimhan, and Griffiths}{Sumers et~al\mbox{.}}{2024}]%
        {cognitive_architectures}
\bibfield{author}{\bibinfo{person}{Theodore~R. Sumers}, \bibinfo{person}{Shunyu Yao}, \bibinfo{person}{Karthik Narasimhan}, {and} \bibinfo{person}{Thomas~L. Griffiths}.} \bibinfo{year}{2024}\natexlab{}.
\newblock \bibinfo{title}{Cognitive Architectures for Language Agents}.
\newblock
\newblock
\showeprint[arxiv]{2309.02427}~[cs.AI]
\urldef\tempurl%
\url{https://arxiv.org/abs/2309.02427}
\showURL{%
\tempurl}


\bibitem[\protect\citeauthoryear{Touvron, Lavril, Izacard, Martinet, Lachaux, Lacroix, Rozière, Goyal, Hambro, Azhar, Rodriguez, Joulin, Grave, and Lample}{Touvron et~al\mbox{.}}{2023}]%
        {touvron2023llama}
\bibfield{author}{\bibinfo{person}{Hugo Touvron}, \bibinfo{person}{Thibaut Lavril}, \bibinfo{person}{Gautier Izacard}, \bibinfo{person}{Xavier Martinet}, \bibinfo{person}{Marie-Anne Lachaux}, \bibinfo{person}{Timothée Lacroix}, \bibinfo{person}{Baptiste Rozière}, \bibinfo{person}{Naman Goyal}, \bibinfo{person}{Eric Hambro}, \bibinfo{person}{Faisal Azhar}, \bibinfo{person}{Aurelien Rodriguez}, \bibinfo{person}{Armand Joulin}, \bibinfo{person}{Edouard Grave}, {and} \bibinfo{person}{Guillaume Lample}.} \bibinfo{year}{2023}\natexlab{}.
\newblock \bibinfo{title}{LLaMA: Open and Efficient Foundation Language Models}.
\newblock
\newblock
\showeprint[arxiv]{2302.13971}~[cs.CL]
\urldef\tempurl%
\url{https://arxiv.org/abs/2302.13971}
\showURL{%
\tempurl}


\bibitem[\protect\citeauthoryear{Wang, Xu, Lan, Hu, Lan, Lee, and Lim}{Wang et~al\mbox{.}}{2023}]%
        {wang2023plan}
\bibfield{author}{\bibinfo{person}{Lei Wang}, \bibinfo{person}{Wanyu Xu}, \bibinfo{person}{Yihuai Lan}, \bibinfo{person}{Zhiqiang Hu}, \bibinfo{person}{Yunshi Lan}, \bibinfo{person}{Roy Ka-Wei Lee}, {and} \bibinfo{person}{Ee-Peng Lim}.} \bibinfo{year}{2023}\natexlab{}.
\newblock \bibinfo{title}{Plan-and-Solve Prompting: Improving Zero-Shot Chain-of-Thought Reasoning by Large Language Models}.
\newblock
\newblock
\showeprint[arxiv]{2305.04091}~[cs.CL]
\urldef\tempurl%
\url{https://arxiv.org/abs/2305.04091}
\showURL{%
\tempurl}


\bibitem[\protect\citeauthoryear{Wei, Bosma, Zhao, Guu, Yu, Lester, Du, Dai, and Le}{Wei et~al\mbox{.}}{2022a}]%
        {wei2022finetuned}
\bibfield{author}{\bibinfo{person}{Jason Wei}, \bibinfo{person}{Maarten Bosma}, \bibinfo{person}{Vincent~Y. Zhao}, \bibinfo{person}{Kelvin Guu}, \bibinfo{person}{Adams~Wei Yu}, \bibinfo{person}{Brian Lester}, \bibinfo{person}{Nan Du}, \bibinfo{person}{Andrew~M. Dai}, {and} \bibinfo{person}{Quoc~V. Le}.} \bibinfo{year}{2022}\natexlab{a}.
\newblock \bibinfo{title}{Finetuned Language Models Are Zero-Shot Learners}.
\newblock
\newblock
\showeprint[arxiv]{2109.01652}~[cs.CL]
\urldef\tempurl%
\url{https://arxiv.org/abs/2109.01652}
\showURL{%
\tempurl}


\bibitem[\protect\citeauthoryear{Wei, Wang, Schuurmans, Bosma, Ichter, Xia, Chi, Le, and Zhou}{Wei et~al\mbox{.}}{2022b}]%
        {wei2022chain}
\bibfield{author}{\bibinfo{person}{Jason Wei}, \bibinfo{person}{Xuezhi Wang}, \bibinfo{person}{Dale Schuurmans}, \bibinfo{person}{Maarten Bosma}, \bibinfo{person}{Brian Ichter}, \bibinfo{person}{Fei Xia}, \bibinfo{person}{Ed~H Chi}, \bibinfo{person}{Quoc Le}, {and} \bibinfo{person}{Denny Zhou}.} \bibinfo{year}{2022}\natexlab{b}.
\newblock \showarticletitle{Chain of thought prompting elicits reasoning in large language models}.
\newblock \bibinfo{journal}{\emph{arXiv preprint arXiv:2201.11903}} (\bibinfo{year}{2022}).
\newblock


\bibitem[\protect\citeauthoryear{Yao, Zhao, Yu, Du, Shafran, Narasimhan, and Cao}{Yao et~al\mbox{.}}{2023}]%
        {yao2022react}
\bibfield{author}{\bibinfo{person}{Shunyu Yao}, \bibinfo{person}{Jeffrey Zhao}, \bibinfo{person}{Dian Yu}, \bibinfo{person}{Nan Du}, \bibinfo{person}{Izhak Shafran}, \bibinfo{person}{Karthik Narasimhan}, {and} \bibinfo{person}{Yuan Cao}.} \bibinfo{year}{2023}\natexlab{}.
\newblock \bibinfo{title}{ReAct: Synergizing Reasoning and Acting in Language Models}.
\newblock
\newblock
\showeprint[arxiv]{2210.03629}~[cs.CL]
\urldef\tempurl%
\url{https://arxiv.org/abs/2210.03629}
\showURL{%
\tempurl}


\bibitem[\protect\citeauthoryear{Zhou, Pujara, Ren, Chen, Cheng, Le, Chi, Zhou, Mishra, and Zheng}{Zhou et~al\mbox{.}}{2024}]%
        {zhou2024selfdiscoverlargelanguagemodels}
\bibfield{author}{\bibinfo{person}{Pei Zhou}, \bibinfo{person}{Jay Pujara}, \bibinfo{person}{Xiang Ren}, \bibinfo{person}{Xinyun Chen}, \bibinfo{person}{Heng-Tze Cheng}, \bibinfo{person}{Quoc~V. Le}, \bibinfo{person}{Ed~H. Chi}, \bibinfo{person}{Denny Zhou}, \bibinfo{person}{Swaroop Mishra}, {and} \bibinfo{person}{Huaixiu~Steven Zheng}.} \bibinfo{year}{2024}\natexlab{}.
\newblock \bibinfo{title}{Self-Discover: Large Language Models Self-Compose Reasoning Structures}.
\newblock
\newblock
\showeprint[arxiv]{2402.03620}~[cs.AI]
\urldef\tempurl%
\url{https://arxiv.org/abs/2402.03620}
\showURL{%
\tempurl}


\end{thebibliography}


\end{document}